# On the Link between Partial Meet, Kernel, and Infra Contraction and its Application to Horn Logic

**Richard Booth**                                                                                             RICHARD.BOOTH@UNI.LU
*Université du Luxembourg*
*Luxembourg*

**Thomas Meyer**                                                                                         TOMMIE.MEYER@MERAKA.ORG.ZA
*Centre for Artificial Intelligence Research*
*University of KwaZulu-Natal and CSIR Meraka Institute*
*South Africa*

**Ivan Varzinczak**                                                                                       IVAN.VARZINCZAK@MERAKA.ORG.ZA
*Centre for Artificial Intelligence Research*
*University of KwaZulu-Natal and CSIR Meraka Institute*
*South Africa*

**Renata Wassermann**                                                                                                 RENATA@IME.USP.BR
*Universidade de São Paulo*
*Brazil*

## Abstract

Standard belief change assumes an underlying logic containing full classical propositional logic. However, there are good reasons for considering belief change in less expressive logics as well. In this paper we build on recent investigations by Delgrande on *contraction* for *Horn logic*. We show that the standard basic form of contraction, *partial meet*, is too strong in the Horn case. This result stands in contrast to Delgrande's conjecture that *orderly maxichoice* is *the* appropriate form of contraction for Horn logic. We then define a more appropriate notion of basic contraction for the Horn case, influenced by the convexity property holding for full propositional logic and which we refer to as *infra contraction*. The main contribution of this work is a result which shows that the construction method for Horn contraction for belief sets based on our infra remainder sets corresponds exactly to Hansson's classical kernel contraction for belief sets, when restricted to Horn logic. This result is obtained via a detour through contraction for belief bases. We prove that kernel contraction for belief bases produces precisely the same results as the belief base version of infra contraction. The use of belief bases to obtain this result provides evidence for the conjecture that Horn belief change is best viewed as a 'hybrid' version of belief set change and belief base change. One of the consequences of the link with base contraction is the provision of a representation result for Horn contraction for belief sets in which a version of the Core-retainment postulate features.

## 1. Introduction

In his seminal paper, Delgrande (2008) has shed some light on the theoretical underpinnings of belief change by weakening a usual assumption in the belief change community, namely that the underlying logical formalism should be at least as strong as (full) classical propositional logic (Gärdenfors, 1988). Delgrande investigated *contraction* functions for *belief sets* (sets of sentences that are closed under logical consequence) restricted to Horn formu-las (1951). Delgrande's main contributions were essentially threefold. Firstly, he showed that the move to Horn logic leads to two different types of contraction functions, referred to as *entailment*-based contraction (*e*-contraction) and *inconsistency*-based contraction (*i*-contraction), which coincide in the full propositional case. Secondly, he showed that Horn contraction for belief sets does not satisfy the controversial *Recovery* postulate, but exhibits some characteristics that are usually associated with the contraction of belief *bases* (arbitrary sets of sentences). And finally, Delgrande made a tentative conjecture that a version of Horn contraction usually referred to as *orderly maxichoice* contraction is *the* appropriate method for contraction in Horn theories.

While Delgrande's partial meet constructions are appropriate choices for contraction in Horn logic, we show that they do not constitute *all the appropriate forms* of Horn contraction. Moreover, as referred to above, although Horn contraction is defined for Horn belief sets, it is related in some ways to contraction for belief bases, an aspect which has not yet been explored properly in the literature.

In this paper we continue the investigation into contraction for Horn logic, and address both the issues mentioned above, as well as others. Focusing on Delgrande's entailment-based contraction, we start by providing a more fine-grained construction for belief set contraction which we refer to in this paper as *infra contraction*. We then bring into the picture a construction method for contraction first introduced by Hansson (1994), known as *kernel contraction*. Although kernel contraction is usually associated with belief base contraction, it can be applied to belief sets as well. Our main contribution is a result which shows that infra contraction corresponds exactly to Hansson's kernel contraction for belief sets, when restricted to Horn logic. In order to prove this, we first take a close look at contraction for belief bases, defining a base version of infra contraction and proving that this construction is equivalent to kernel contraction for Horn belief bases. Since Horn belief sets are not closed under classical logical consequence, they can be seen as a 'hybrid' between belief sets and belief bases. This justifies the use of belief bases to obtain results for belief set Horn contraction.

Horn logic has found extensive use in Artificial Intelligence, in particular in logic programming, truth maintenance systems, and deductive databases. This explains, in part, our interest in belief change for Horn logic. (Despite our interest in Horn formulas, it is worth noting that in this work we do not consider logic programming explicitly and we do not use negation as failure at all.) Another reason for focusing on this topic is because of its application to debugging and repairing ontologies in description logics (Baader, Calvanese, McGuinness, Nardi, & Patel-Schneider, 2007). In particular, Horn logic can be seen as the backbone of the $\mathcal{EL}$ family of description logics (Baader, Brandt, & Lutz, 2005), and therefore a proper understanding of belief change for Horn logic is important for finding solutions to similar problems expressible in the $\mathcal{EL}$ family.

The remainder of the present paper is organized as follows: After some logical preliminaries (Section 2), we give the background on belief set contraction (Section 3) and on belief base contraction (Section 4) that is necessary for the core section of the paper (Section 5). There we prove that kernel contraction and infra contraction are equivalent on the level of belief bases. This enables us to prove that kernel contraction and infra contraction are equivalent on the Horn belief set level as well. From this we are led to provide a characterization of infra contraction for Horn belief sets in which a version of the Core-retainment postulate (Hansson, 1994) for bases features. This provides even more evidence for the 'hybrid' aspect of Horn belief change. All these results are stated for entailment-based contraction (*e*-contraction). In Section 6, where we discuss related work, we also mention similar results for Delgrande's *i*-contraction and another relevant type of Horn contraction, namely Booth et al's (2009) *package* contraction. We conclude with a summary of our contributions as well as a discussion on future directions of investigation. Proofs of all new results can be found in Appendix A.

## 2. Preliminaries

We work in a finitely generated propositional language over a set of propositional *atoms* $\mathfrak{P}$, together with the distinguished atom $\top$ (*true*), and with the standard model-theoretic semantics. Atoms are denoted by $p, q, \ldots$, possibly with subscripts. The *formulas* of our language are denoted by $\varphi, \psi, \ldots$ Those are recursively defined as follows:

$$\varphi ::= p \mid \top \mid \neg\varphi \mid \varphi \wedge \varphi$$

All the other connectives ($\vee, \rightarrow, \leftrightarrow, \ldots$) and the special atom $\bot$ (*false*) are defined in terms of $\neg$ and $\wedge$ in the usual way. With $\mathcal{L}_\mathsf{P}$ we denote the set of all formulas of the language.

Classical logical consequence and logical equivalence are denoted by $\models$ and $\equiv$ respectively. For $X \subseteq \mathcal{L}_\mathsf{P}$, the set of sentences logically entailed by $X$ is denoted by $Cn(X)$. A *belief set* is a logically closed set, i.e., for a belief set $K$, $K = Cn(K)$. We usually denote belief sets by $K$, possibly decorated by primes. $\mathscr{P}(X)$ denotes the power set (set of all subsets) of $X$. When displaying belief sets we sometimes follow the convention of displaying one representative of each equivalence class modulo logical equivalence, and dropping the representative for the tautologies. As an example, for $\mathcal{L}_\mathsf{P}$ generated by the two atoms $p$ and $q$, the set $Cn(\{p\})$ can be represented as $\{p, p \vee q, p \vee \neg q\}$.

A *Horn clause* is a sentence of the form $p_1 \wedge p_2 \wedge \ldots \wedge p_n \rightarrow q$ where $n \geq 0$, $p_i, q \in \mathfrak{P}$ for $1 \leq i \leq n$ (recall that the $p_i$s and $q$ may be one of $\bot$ or $\top$ as well). If $n = 0$ we write $q$ instead of $\rightarrow q$. A *Horn formula* is a conjunction of Horn clauses. A *Horn set* is a set of Horn formulas.

Given a propositional language $\mathcal{L}_\mathsf{P}$, the Horn language $\mathcal{L}_\mathsf{H}$ generated from $\mathcal{L}_\mathsf{P}$ is simply the set of Horn formulas occurring in $\mathcal{L}_\mathsf{P}$. The Horn logic obtained from $\mathcal{L}_\mathsf{H}$ has the same semantics as the propositional logic obtained from $\mathcal{L}_\mathsf{P}$, but just restricted to Horn formulas. Hence, we have that $\models, \equiv$, and all other related notions are defined relative to the logic we are working in. We use $\models_{\overline{\mathsf{PL}}}$ and $Cn_{\mathsf{PL}}(.)$ to denote classical entailment and consequence for propositional logic. For Horn logic, we define $Cn_{\mathsf{HL}}(.)$ as follows: $Cn_{\mathsf{HL}}(X) =_{\text{def}} Cn_{\mathsf{PL}}(X) \cap \mathcal{L}_\mathsf{H}$ for $X \subseteq \mathcal{L}_\mathsf{H}$. And we define $\models_{\overline{\mathsf{HL}}}$ as follows: For $X \subseteq \mathcal{L}_\mathsf{H}$ and $\varphi \in \mathcal{L}_\mathsf{H}$, $X \models_{\overline{\mathsf{HL}}} \varphi$ if and only if $X \models_{\overline{\mathsf{PL}}} \varphi$.

The consequence operator $Cn_{\mathsf{HL}}(.)$ is a *Tarskian consequence operator* in the sense that it satisfies the following properties for all Horn sets $X, X'$:

- $X \subseteq Cn_{\mathsf{HL}}(X)$ (Inclusion)

- $Cn_{\mathsf{HL}}(X) = Cn_{\mathsf{HL}}(Cn_{\mathsf{HL}}(X))$ (Idempotency)

- If $X \subseteq X'$, then $Cn_{\mathsf{HL}}(X) \subseteq Cn_{\mathsf{HL}}(X')$ (Monotonicity)

A *Horn belief set*, usually denoted by $H$ (possibly with primes), is a Horn set closed under the operator $Cn_{\mathsf{HL}}(.)$, i.e., $H = Cn_{\mathsf{HL}}(H)$. We shall dispense with such subscripts whenever the context makes it clear which logic we are dealing with.

In the AI tradition, a given set of formulas of the underlying logical language is called a *knowledge base*, or simply a set of *beliefs*. Belief change deals with situations in which an agent has to modify its beliefs about the world, usually due to new or previously unknown incoming information, also represented as formulas of the language. Common operations of interest in belief change are the *expansion* (Gärdenfors, 1988) of an agent's current beliefs $X$ by a given formula $\varphi$ (usually denoted as $X + \varphi$), where the basic idea is to add $\varphi$ regardless of the consequences, and the *revision* (Gärdenfors, 1988) of its current beliefs by $\varphi$ (denoted as $X \star \varphi$), where the intuition is to incorporate $\varphi$ into the current beliefs in some way while ensuring consistency of the resulting theory at the same time. Perhaps the most basic operation in belief change is that of *contraction* (Alchourrón, Gärdenfors, & Makinson, 1985; Gärdenfors, 1988), which is intended to represent situations in which an agent has to give up $\varphi$ from its current stock of beliefs (denoted as $X - \varphi$). Indeed the revision operation can be defined in terms of contraction and simple expansion via the Levi identity (Levi, 1977). Therefore, contraction is the focus of the present paper and in what follows we shall investigate it in detail. Throughout Sections 3 and 4 we assume we work in the language of full propositional logic $\mathcal{L}_\mathsf{P}$.

## 3. Belief Set Contraction

We commence with a discussion on *belief set* contraction, where the aim is to describe contraction on the *knowledge level* (Gärdenfors, 1988), i.e., independently of how beliefs are represented syntactically. Thus, contraction is defined only for *belief sets*.

**Definition 1 (Belief Set Contraction)** *A* belief set contraction $-$ *for a belief set $K$ is a function from $\mathcal{L}_\mathsf{P}$ to $\mathscr{P}(\mathcal{L}_\mathsf{P})$.*

By the principle of categorical matching (Gärdenfors & Rott, 1995) the contraction of a belief set by a sentence is expected to yield a new belief set.

One of the standard approaches for constructing belief contraction operators is based on the notion of a *remainder set* of a set $K$ with respect to a formula $\varphi$: a maximal subset of $K$ not entailing $\varphi$ (Alchourrón et al., 1985). Here we define this for belief sets, but it has been known in the literature that its basic principle can be applied to belief bases as well (we shall recall this in Section 4).

**Definition 2 (Remainder Sets)** *Given a belief set $K$ and a formula $\varphi \in \mathcal{L}_\mathsf{P}$, $X \in K \perp \varphi$ if and only if*

- $X \subseteq K$;
- $X \not\models \varphi$; *and*
- *For every $X'$ such that $X \subset X' \subseteq K$, $X' \models \varphi$.*

*We call the elements of $K \perp \varphi$ the* remainder sets *of $K$ with respect to $\varphi$.*

It is easy to verify that $K \perp \varphi = \emptyset$ if and only if $\models \varphi$ (Gärdenfors, 1988).

Since there is no unique method for choosing between possibly different remainder sets, there is a presupposition of the existence of a suitable *selection function* for doing so:

**Definition 3 (Selection Functions)** *A selection function $\gamma$ for a set $K$ is a (partial) function from $\mathscr{P}(\mathscr{P}(\mathcal{L}_\mathsf{P}))$ to $\mathscr{P}(\mathscr{P}(\mathcal{L}_\mathsf{P}))$ such that*

- $\gamma(K \perp \varphi) = \{K\}$ *if $K \perp \varphi = \emptyset$; and*

- $\emptyset \subset \gamma(K \perp \varphi) \subseteq K \perp \varphi$ *otherwise.*

Selection functions provide a mechanism for identifying the remainder sets judged to be most appropriate. The resulting contraction is obtained by taking the intersection of the chosen remainder sets.

**Definition 4 (Partial Meet Contraction)** *For a selection function $\gamma$, the belief set contraction operator $-_\gamma$ generated by $\gamma$ and defined as $K -_\gamma \varphi =_{def} \bigcap \gamma(K \perp \varphi)$ is a* partial meet *contraction.*

Two subclasses of belief set partial meet deserve special mention:

**Definition 5 (Maxichoice and Full Meet)** *Given a selection function $\gamma$, $-_\gamma$ is a* maxichoice *contraction if and only if $\gamma(K \perp \varphi)$ is always a singleton set. It is a* full meet *contraction if and only if $\gamma(K \perp \varphi) = K \perp \varphi$ whenever $K \perp \varphi \neq \emptyset$.*

It is worth mentioning that belief set full meet contraction is unique, while belief set maxichoice contraction usually is not.

*Kernel contraction* was introduced by Hansson (1994) as a generalization of *safe contraction* (Alchourrón & Makinson, 1985). Instead of looking at *maximal* subsets *not* implying a given formula, kernel operations are based on *minimal* subsets that *do* imply it.

**Definition 6 (Kernel)** *For a belief set $K$, $X \in K \perp\!\!\!\perp \varphi$ if and only if*

- $X \subseteq K$;

- $X \models \varphi$; *and*

- *For every $X'$ such that $X' \subset X$, $X' \not\models \varphi$.*

$K \perp\!\!\!\perp \varphi$ is called the kernel set *of $K$ with respect to $\varphi$ and the elements of $K \perp\!\!\!\perp \varphi$ are called the $\varphi$-kernels of the belief set $K$.*

The result of a kernel contraction is obtained by removing at least one element from every (non-empty) $\varphi$-kernel of $K$, using an *incision function*.

**Definition 7 (Incision Functions)** *An incision function $\sigma$ for a set $K$ is a function from the set of kernel sets of $K$ to $\mathscr{P}(\mathcal{L}_\mathsf{P})$ such that*

- $\sigma(K \perp\!\!\!\perp \varphi) \subseteq \bigcup (K \perp\!\!\!\perp \varphi)$; and

- If $\emptyset \neq X \in K \perp\!\!\!\perp \varphi$, then $X \cap \sigma(K \perp\!\!\!\perp \varphi) \neq \emptyset$.

Given a belief set $K$ and an incision function $\sigma$ for $K$, ideally we would want the kernel contraction $-_\sigma$ for $K$ generated by $\sigma$ be defined as:

$$K -_\sigma \varphi =_{\text{def}} K \setminus \sigma(K \perp\!\!\!\perp \varphi) \tag{1}$$

It turns out that for a belief set $K$, the contraction operator in (1) does not respect the principle of categorical matching. That is because, for a belief set as input, the kernel contraction $-_\sigma$ from (1) above does not necessarily produce a belief set as a result: $K -_\sigma \varphi$ is not in general closed under logical consequence, as shown by the following example.

**Example 1** *Let $K = Cn(\{p \to q, q \to r\})$ and assume that we want to contract $p \to r$ from $K$. We have $K \perp\!\!\!\perp (p \to r) = \{\{p \to r\}, \{p \to q, q \to r\}, \{p \to q, p \wedge q \to r\}\}$. For the incision function $\sigma$ defined such that $\sigma(K \perp\!\!\!\perp (p \to r)) = \{p \to r, p \to q, p \wedge q \to r\}$, we have that applying the operator in (1) gives us $K'$ such that $K' \models p \wedge q \to r$, but obviously $p \wedge q \to r \notin K'$.*

Of course, it is possible to ensure that one obtains a belief set by closing the result obtained from the operator above under logical consequence.

**Definition 8 (Belief Set Kernel Contraction)** *Given a belief set $K$ and an incision function $\sigma$ for $K$, the belief set contraction operator $\approx_\sigma$ for $K$ generated by $\sigma$ and defined as $K \approx_\sigma \varphi =_{def} Cn(K -_\sigma \varphi)$ is a* belief set kernel contraction.

We shall see in Section 4 that belief set kernel contraction is closely related to a version of belief base contraction referred to as *saturated base kernel contraction* (Hansson, 1999).

Belief set contraction defined in terms of partial meet contraction corresponds exactly to what is perhaps the best-known approach to belief change: the so-called AGM approach (Alchourrón et al., 1985). AGM requires that (basic) belief set contraction be characterized by the following set of postulates:

$(K-1)$ $K - \varphi = Cn(K - \varphi)$ (Closure)

$(K-2)$ $K - \varphi \subseteq K$ (Inclusion)

$(K-3)$ If $\varphi \notin K$, then $K - \varphi = K$ (Vacuity)

$(K-4)$ If $\not\models \varphi$, then $\varphi \notin K - \varphi$ (Success)

$(K-5)$ If $\varphi \equiv \psi$, then $K - \varphi = K - \psi$ (Extensionality)

$(K-6)$ If $\varphi \in K$, then $Cn((K - \varphi) \cup \{\varphi\}) = K$ (Recovery)

Full AGM contraction involves two extended postulates in addition to the basic postulates given above. We shall not elaborate in detail on the intuition of these postulates. For that we refer the reader to the book by Gärdenfors (1988) or the handbook by Hansson (1999).

Alchourrón et al. (1985) have shown that these postulates characterize belief set partial meet contraction exactly, as shown by the following result:

**Theorem 1 (Alchourrón et al., 1985)** *Every belief set partial meet contraction satisfies Postulates $(K-1)$–$(K-6)$. Conversely, every belief set contraction which satisfies Postulates $(K-1)$–$(K-6)$ is a belief set partial meet contraction.*

Hansson (1994) has shown that Postulates $(K-1)$–$(K-6)$ also characterize kernel contraction for belief sets:

**Theorem 2 (Hansson, 1994)** *Every belief set kernel contraction satisfies the Postulates $(K-1)$–$(K-6)$. Conversely, every belief set contraction which satisfies Postulates $(K-1)$–$(K-6)$ is a belief set kernel contraction.*

We conclude this section with an important new observation whose usefulness will become apparent in Section 5.

**Theorem 3 (Convexity)** *Let $K$ be a belief set, let $-_{mc}$ be a (belief set) maxichoice contraction, and let $-_{fm}$ denote (belief set) full meet contraction. For every $\varphi \in \mathcal{L}_\mathsf{P}$ and every belief set $X_\varphi$ such that $K -_{fm} \varphi \subseteq X_\varphi \subseteq K -_{mc} \varphi$, there is a (belief set) partial meet contraction $-_{pm}$ such that $K -_{pm} \varphi = X_\varphi$.*

This result shows that every belief set between the results obtained from full meet contraction and some maxichoice contraction can also be obtained from some partial meet contraction. So, it is possible to define a version of belief set contraction based on such sets.

**Definition 9 (Infra Remainder Sets)** *Given a formula $\varphi \in \mathcal{L}_\mathsf{P}$, and belief sets $K$ and $K'$, $K' \in K \downarrow \varphi$ if and only if there is some $K'' \in K \bot \varphi$ such that $(\bigcap K \bot \varphi) \subseteq K' \subseteq K''$. We refer to the elements of $K \downarrow \varphi$ as the* infra remainder sets *of $K$ with respect to $\varphi$.*

It is easy to see that $K \downarrow \varphi = \emptyset$ if and only if $\models \varphi$. Note that by definition the set of infra remainder sets of a belief set $K$ are required to be belief sets themselves.

All remainder sets with respect to a formula $\varphi$ are also infra remainder sets with respect to $\varphi$, and so is the intersection of any set of remainder sets with respect to $\varphi$. Indeed, the intersection of any set of infra remainder sets with respect to $\varphi$ is also an infra remainder set with respect to $\varphi$. So the set of infra remainder sets with respect to $\varphi$ contains *all* belief sets between some remainder set with respect to $\varphi$ and the intersection of all remainder sets with respect to $\varphi$. This explains why infra contraction below is not defined as the intersection of infra remainder sets (cf. Definition 4).

**Definition 10 (Infra Contraction)** *Let $K$ be a belief set. An infra selection function $\tau$ for $K$ is a (partial) function from $\mathscr{P}(\mathscr{P}(\mathcal{L}_\mathsf{P}))$ to $\mathscr{P}(\mathcal{L}_\mathsf{P})$ such that $\tau(K \downarrow \varphi) = K$ whenever $K \downarrow \varphi = \emptyset$, and $\tau(K \downarrow \varphi) \in K \downarrow \varphi$ otherwise. A belief set contraction operator $-_\tau$ is a* belief set infra contraction *if and only if $K -_\tau \varphi = \tau(K \downarrow \varphi)$.*

In a sense, then, infra contraction is closer to maxichoice contraction than to partial meet contraction, since it chooses just a single element from the set of infra remainder sets, much like maxichoice contraction chooses just a single element from the set of remainder sets.

One immediate consequence of Theorem 3 and Definition 10 is the following:

**Corollary 1** *Infra contraction and partial meet contraction coincide for belief sets.*

## 4. Belief Base Contraction

Now we turn our attention to belief *base* contraction, where an agent's beliefs are represented as a (usually finite) set of sentences not necessarily closed under logical consequence, also known as a *base*. We usually denote bases by $B$, possibly decorated with primes.

**Definition 11 (Belief Base Contraction)** *A base contraction $-$ for a base $B$ is a function from $\mathcal{L}_\mathsf{P}$ to $\mathscr{P}(\mathcal{L}_\mathsf{P})$.*

Intuitively the idea is that, for a fixed belief base $B$, contraction of a formula $\varphi$ from $B$ produces a new base $B - \varphi$.

Given the construction methods for belief set contraction already discussed in Section 3, two obvious ways to define belief base contraction are to consider both partial meet contraction and kernel contraction for bases. Below we present both cases.

All definitions required for partial meet in the base case are analogous to those given in the belief set case in Section 3. For the reader's convenience, we state them explicitly here:

**Definition 12 (Base Remainder Sets)** *Given a base $B$ and a formula $\varphi$, $X \in B \bot \varphi$ if and only if*

- $X \subseteq B$;
- $X \not\models \varphi$; and
- *For every $X'$ such that $X \subset X' \subseteq B$, $X' \models \varphi$.*

*We call the elements of $B \bot \varphi$ the* base remainder sets *of $B$ with respect to $\varphi$.*

Similarly to the belief set case, it is easy to verify that $B \bot \varphi = \emptyset$ if and only if $\models \varphi$.

**Definition 13 (Selection Functions)** *A selection function $\gamma$ for a base $B$ is a (partial) function from $\mathscr{P}(\mathscr{P}(\mathcal{L}_\mathsf{P}))$ to $\mathscr{P}(\mathscr{P}(\mathcal{L}_\mathsf{P}))$ such that*

- $\gamma(B \bot \varphi) = \{B\}$ *if $B \bot \varphi = \emptyset$; and*
- $\emptyset \subset \gamma(B \bot \varphi) \subseteq B \bot \varphi$ *otherwise.*

**Definition 14 (Base Partial Meet Contraction)** *For a selection function $\gamma$, the belief base contraction operator $-_\gamma$ generated by $\gamma$ and defined as $B -_\gamma \varphi =_{def} \bigcap \gamma(B \bot \varphi)$ is a base partial meet contraction.*

**Definition 15 (Base Maxichoice and Full Meet)** *Given a selection function $\gamma$, $-_\gamma$ is a* base maxichoice *contraction if and only if $\gamma(B\bot\varphi)$ is always a singleton set. It is a* base full meet *contraction if and only if $\gamma(B\bot\varphi) = B\bot\varphi$ whenever $B\bot\varphi \neq \emptyset$.*

As in the belief set case, it can be checked that belief base full meet contraction is unique, while base maxichoice contraction usually is not.

Hansson (1992) showed that belief base partial meet contraction can be characterized by the following postulates:

$(B-1)$ If $\not\models \varphi$, then $B - \varphi \not\models \varphi$ \hfill (Success)

$(B-2)$ $B - \varphi \subseteq B$ \hfill (Inclusion)

$(B-3)$ If $B' \models \varphi$ if and only if $B' \models \psi$ for all $B' \subseteq B$, then $B - \varphi = B - \psi$ \hfill (Uniformity)

$(B-4)$ If $\psi \in (B \setminus (B - \varphi))$, then there is a $B'$ such that
$(B - \varphi) \subseteq B' \subseteq B$ and $B' \not\models \varphi$, but $B' \cup \{\psi\} \models \varphi$ \hfill (Relevance)

Again, we shall not elaborate in detail on the intuition of these postulates. For that we refer the reader to the handbook by Hansson (1999).

**Theorem 4 (Hansson, 1992)** *Every base partial meet contraction operator satisfies Postulates $(B-1)$–$(B-4)$. Conversely, every base contraction which satisfies $(B-1)$–$(B-4)$ is a base partial meet contraction.*

One question that arises is whether the following belief *base* version of the convexity principle from Theorem 3 holds:

**(Base Convexity)** For a belief base $B$, let $-_{mc}$ be a base maxichoice contraction, and let $-_{fm}$ denote base full meet contraction. For every set $X$ and $\varphi$ such that $B -_{fm} \varphi \subseteq X \subseteq B -_{mc} \varphi$, there is a base partial meet contraction $-_{pm}$ such that $B -_{pm} \varphi = X$.

As in the belief set case, this principle simply states that every set between the results obtained from base full meet contraction and some base maxichoice contraction can be obtained from some belief base partial meet contraction. The following example shows that it does *not* hold.

**Example 2** *Let $B = \{p \to q, q \to r, p \wedge q \to r, p \wedge r \to q\}$ and consider contraction by $p \to r$. It is easily verified that base maxichoice gives either $B - (p \to r) = B' = \{p \to q, p \wedge r \to q\}$ or $B - (p \to r) = B'' = \{q \to r, p \wedge q \to r, p \wedge r \to q\}$. Therefore the only other result obtained from base partial meet contraction is that which is provided by base full meet contraction: $B - (p \to r) = B''' = \{p \wedge r \to q\}$. But observe that even though it is the case that $B''' \subseteq X \subseteq B''$ where $X = \{p \wedge q \to r, p \wedge r \to q\}$, $X$ is not equal to any of $B'$, $B''$, or $B'''$.*

We shall come back to this issue at the end of the section.

Below we give the definitions for kernel contraction in the belief base case. Again, they are analogous to those given in the belief set case in the previous section, but for the reader's convenience we state them explicitly here.

**Definition 16 (Base Kernels)** *For a belief base $B$ and a formula $\varphi$, $X \in B \bot\!\!\!\bot \varphi$ if and only if*

- $X \subseteq B$;
- $X \models \varphi$; and
- *For every $X'$ such that $X' \subset X$, $X' \not\models \varphi$.*

$B \bot\!\!\!\bot \varphi$ *is called the* kernel set *of $B$ with respect to $\varphi$, and the elements of $B \bot\!\!\!\bot \varphi$ are called the $\varphi$-kernels of $B$.*

The result of a base kernel contraction is obtained by removing at least one element from every (non-empty) $\varphi$-kernel of $B$, using an incision function.

**Definition 17 (Incision Functions for Bases)** *An* incision function $\sigma$ *for a base $B$ is a function from the set of kernel sets of $B$ to $\mathscr{P}(\mathcal{L}_\mathsf{P})$ such that*

- $\sigma(B \bot\!\!\!\bot \varphi) \subseteq \bigcup(B \bot\!\!\!\bot \varphi)$; *and*
- *If $\emptyset \neq X \in B \bot\!\!\!\bot \varphi$, then $X \cap \sigma(B \bot\!\!\!\bot \varphi) \neq \emptyset$.*

*We refer to an incision function $\sigma$ as* minimal *if and only if for every $\varphi$, and every incision function $\sigma'$, $\sigma(B \bot\!\!\!\bot \varphi) \subseteq \sigma'(B \bot\!\!\!\bot \varphi)$. The (unique)* maximum *incision function $\sigma$ is defined as follows: for every $\varphi$, $\sigma(B \bot\!\!\!\bot \varphi) = \bigcup(B \bot\!\!\!\bot \varphi)$.*

It is worthy of mention that minimal incision functions always exist.

**Definition 18 (Base Kernel Contraction)** *Given an incision function $\sigma$ for a base $B$, the* base kernel contraction $-_\sigma$ *for $B$ generated by $\sigma$ is defined as: $B -_\sigma \varphi = B \setminus \sigma(B \bot\!\!\!\bot \varphi)$.*

Base kernel contraction can be characterized by the same postulates as base partial meet contraction, except that Relevance is replaced by the *Core-retainment* postulate below:

$(B-5)$ If $\psi \in (B \setminus (B - \varphi))$, then
there is some $B' \subseteq B$ such that $B' \not\models \varphi$ but $B' \cup \{\psi\} \models \varphi$ \qquad (Core-retainment)

**Theorem 5 (Hansson, 1994)** *Every base kernel contraction satisfies Postulates $(B-1)$–$(B-3)$ and $(B-5)$. Conversely, every base contraction which satisfies Postulates $(B-1)$–$(B-3)$ and $(B-5)$ is a base kernel contraction.*

Clearly, Core-retainment is slightly weaker than Relevance. And indeed, it thus follows that all base partial meet contractions are base kernel contractions, but as the following example from Hansson (1999) shows, some base kernel contractions are not base partial meet contractions.

**Example 3** *Let $B = \{p, p \vee q, p \leftrightarrow q\}$. Then $B \bot\!\!\!\bot (p \wedge q) = \{\{p, p \leftrightarrow q\}, \{p \vee q, p \leftrightarrow q\}\}$. So there is an incision function $\sigma$ for $B$ such that $\sigma(B \bot\!\!\!\bot (p \wedge q)) = \{p \vee q, p \leftrightarrow q\}$, and then $B -_\sigma (p \wedge q) = \{p\}$. On the other hand, $B \bot (p \wedge q) = \{\{p, p \vee q\}, \{p \leftrightarrow q\}\}$, from which it follows that base partial meet contraction $B - (p \wedge q)$ yields either $\{p, p \vee q\}$, or $\{p \leftrightarrow q\}$, or $\{p, p \vee q\} \cap \{p \leftrightarrow q\} = \emptyset$, none of which are equal to $B -_\sigma (p \wedge q) = \{p\}$.*

As we briefly pointed out after Definition 8, our definition of kernel contraction for belief sets is closely related to a version of belief base contraction that Hansson (1999) refers to as saturated base kernel contraction:

**Definition 19 (Saturated Base Kernel Contraction)** *Given a belief base $B$ and an incision function $\sigma$ for $B$, the base contraction $\approx_\sigma$ for $B$ generated by $\sigma$ and defined as $B \approx_\sigma \varphi =_{def} B \cap Cn(B -_\sigma \varphi)$ is a* saturated base kernel contraction.

It is easily shown (Hansson, 1994) that when the set $B$ in the definition for saturated base kernel contraction is a belief set, the two notions coincide.

**Observation 1** *For a belief set $K$ and an incision function $\sigma$ for $K$, the saturated base kernel contraction for $\sigma$ and the belief set kernel contraction for $\sigma$ are identical.*

While saturated base kernel contractions are not in general base partial meet contractions (Hansson, 1999, p. 91), this distinction disappears when considering belief sets only:

**Theorem 6 (Hansson, 1994)** *Let $K$ be a belief set. A belief set contraction $-$ is a saturated base kernel contraction if and only if it is a belief set partial meet contraction.*

And as a result of Observation 1 and Theorem 6 we immediately have the following corollary:

**Corollary 2** *Let $K$ be a belief set. A belief set contraction $-$ is a belief set kernel contraction if and only if it is a belief set partial meet contraction.*

Thanks to Theorem 3 (Convexity) we can extend Corollary 2 to show that kernel contraction, partial meet contraction, and infra contraction all coincide for belief sets.

**Corollary 3** *Let $K$ be a belief set. A belief set contraction $-$ is a belief set kernel contraction if and only if it is a belief set partial meet contraction if and only if it is a belief set infra contraction.*

So far we have considered remainder sets for bases and kernel sets for bases, but not infra remainder sets for bases (cf. end of Section 3). We conclude this section with a discussion on this commencing with the definition of *base infra remainder sets*.

**Definition 20 (Base Infra Remainder Sets)** *Given a formula $\varphi$, and bases $B$ and $B'$, $B' \in B \downarrow \varphi$ if and only if there is some $B'' \in B \perp \varphi$ such that $(\bigcap B \perp \varphi) \subseteq B' \subseteq B''$. We refer to the elements of $B \downarrow \varphi$ as the* base infra remainder sets *of $B$ with respect to $\varphi$.*

Observe that the definition of base infra remainder sets is the same as for infra remainder sets, differing only in that it deals with belief bases and not belief sets. Note in particular that the elements are not required to be belief sets themselves. Base infra remainder sets can be used to define a form of base contraction in a way that is similar to the definition of (belief set) infra contraction (cf. Definition 10).

**Definition 21 (Base Infra Contraction)** *A base infra selection function $\tau$ is a (partial) function from $\mathscr{P}(\mathscr{P}(\mathcal{L}_\mathsf{P}))$ to $\mathscr{P}(\mathcal{L}_\mathsf{P})$ such that $\tau(B \downarrow \varphi) = B$ whenever $B \downarrow \varphi = \emptyset$, and $\tau(B \downarrow \varphi) \in B \downarrow \varphi$ otherwise. A belief base contraction operator $-_\tau$ generated by $\tau$ and defined as $B -_\tau \varphi =_{def} \tau(B \downarrow \varphi)$ is a base infra contraction.*

A natural question to ask is how base infra contraction compares with base partial meet contraction and base kernel contraction. The following result, which plays a central role in this paper, shows that base infra contraction corresponds exactly to base kernel contraction.

**Theorem 7** *A base contraction for a belief base $B$ is a belief base kernel contraction for $B$ if and only if it is a base infra contraction for $B$.*

From Example 3 we know that base kernel contraction is more general than base partial meet contraction: every base partial meet contraction is also a base kernel contraction, but the converse does not hold. From Theorem 7 the following result thus follows.

**Observation 2** *Base infra contraction is more general than base partial meet contraction.*

Theorem 7 has a number of other interesting consequences as well. On a philosophical note, it provides evidence for the contention that the kernel contraction approach is more appropriate than the partial meet approach. As mentioned earlier, it is well-known that kernel contraction and partial meet contraction coincide for belief sets, but differ for belief bases (kernel contraction being more general than partial meet contraction in this case). The importance of Theorem 7 is that it shows a new and seemingly different approach to contraction (infra contraction) to be identical to kernel contraction for both belief sets and belief bases, but different from partial meet contraction for belief bases, thereby tilting the scales of evidence towards kernel contraction. As we shall see in the next section, Theorem 7 is also instrumental in "lifting" this result to the level of Horn belief sets.

## 5. Horn Belief Set Contraction

In the previous sections we recalled several results from the belief change literature and stated new ones, connecting different constructions to sets of rationality postulates. It is important to note that although most of the examples in the literature assume that the underlying logic contains classical propositional logic, the results are usually valid for a broader set of languages. This has been discussed in different contexts by Hansson and Wassermann (2002), Flouris et al. (2006), Ribeiro and Wassermann (2009a, 2009b, 2010), Varzinczak (2008, 2010), and Wassermann (2011), among others. For belief bases, it was shown by Hansson and Wassermann (2002) that in order to keep the representation theorems for partial meet and kernel contraction, the logic only needs to be compact and monotonic. This means that the results transfer directly to Horn logics. However, for belief sets the representation theorems demand more restrictions of the logic. As shown by Flouris et al. (2006), a contraction operation satisfying the basic AGM postulates exists only if the logic is *decomposable*.

**Definition 22 (Flouris et al., 2006)** *A logic is called* decomposable *if and only if for all sets of formulas $X$, $X'$, such that $Cn(\emptyset) \subset Cn(X') \subset Cn(X)$, there exists a set of formulas $X''$ such that $Cn(X'') \subset Cn(X)$ and $Cn(X) = Cn(X' \cup X'')$.*

It was shown by Ribeiro (2010) that Horn logic is not decomposable. Therefore, new constructions and sets of postulates are needed to deal with contraction of Horn belief sets.

Delgrande (2008) investigated two distinct classes of contraction functions for Horn belief sets: *entailment*-based contraction (*e*-contraction), for removing an unwanted consequence; and *inconsistency*-based contraction (*i*-contraction), for removing formulas leading to inconsistency; while Booth et al. (2009) subsequently extended the work of Delgrande by providing more fine-grained versions of his constructions. Our focus in this paper is on *e*-contraction, although Delgrande, as well as Booth et al., also consider *i*-contraction.

Recall that we use $H$, sometimes decorated with primes, to denote a Horn belief set.

**Definition 23 (*e*-Contraction)** *An e-contraction $-$ for a Horn belief set $H$ is a function from $\mathcal{L}_\mathsf{H}$ to $\mathscr{P}(\mathcal{L}_\mathsf{H})$.*

Delgrande's method of construction for *e*-contraction is in terms of partial meet contraction. The definitions of remainder sets (Definition 2), selection functions (Definition 3), partial meet contraction (Definition 4), as well as maxichoice and full meet contraction (Definition 5) all carry over for *e*-contraction, with the set $K$ in each case being replaced by a Horn belief set $H$, and we refer to these as *e*-remainder sets (denoted by $H\perp_e\varphi$), *e*-selection functions, partial meet *e*-contraction, maxichoice *e*-contraction and full meet *e*-contraction respectively (we leave out the reference to the term "Horn", since there is no room for ambiguity here). As in the full propositional case, it is easy to verify that all *e*-remainder sets are also Horn belief sets, and that all partial meet *e*-contractions (and therefore the maxichoice *e*-contractions, as well as full meet *e*-contraction) produce Horn belief sets.

Although Delgrande defines and discusses partial meet *e*-contraction, he argues that maxichoice *e*-contraction is *the* appropriate approach for *e*-contraction. To be more precise, the version of maxichoice *e*-contraction that Delgrande advocates is actually a restricted version of maxichoice *e*-contraction that we refer to as *orderly* maxichoice *e*-contraction. (This is just the Horn case of the maxichoice contraction operators satisfying all the AGM postulates, including the supplementary ones). Whereas (ordinary) maxichoice *e*-contraction is constructed by setting, for every $\varphi \in \mathcal{L}_\mathsf{H}$, $H - \varphi$ equal to *any* element of $H\perp\varphi$, orderly maxichoice *e*-contraction is more systematic in nature. Associated with every orderly maxichoice *e*-contraction is a *linear* order $\leq$ on $\bigcup_{\varphi \in \mathcal{L}_\mathsf{H}}(H\perp\varphi)$, the union of all *e*-remainder sets of $H$ with respect to all sentences $\varphi \in \mathcal{L}_\mathsf{H}$. Intuitively, the higher up in the linear order, the more plausible an element of $\bigcup_{\varphi \in \mathcal{L}_\mathsf{H}}(H\perp\varphi)$ is intended to be. The orderly maxichoice *e*-contraction generated from $\leq$ is then obtained as follows: for every $\varphi \in \mathcal{L}_\mathsf{H}$, $K - \varphi$ is selected to be the most plausible *e*-remainder set of $H$ with respect to $\varphi$ (the one that is the highest up in the order $\leq$).

Here we argue that although all partial meet *e*-contractions are appropriate choices for *e*-contraction, they do not make up the set of *all* appropriate *e*-contractions. In other words, Delgrande's approach is not 'complete'. The argument for appropriate *e*-contractions other than partial meet *e*-contraction is based on the observation that the convexity result for full propositional logic in Theorem 3 does not hold for Horn logic.

**Example 4** *Let $H = Cn_\mathsf{HL}(\{p \to q, q \to r\})$. For the e-contraction of $p \to r$ from $H$, maxichoice yields either $H^1_{mc} = Cn_\mathsf{HL}(\{p \to q\})$ or $H^2_{mc} = Cn_\mathsf{HL}(\{q \to r, p \land r \to q\})$, whereas full* meet yields $H_{fm} = Cn_{\mathsf{HL}}(\{p \wedge r \to q\})$. *These are the only three partial meet e-contractions. Now consider the Horn belief set* $H' = Cn_{\mathsf{HL}}(\{p \wedge q \to r, p \wedge r \to q\})$. *It is clear that* $H_{fm} \subseteq H' \subseteq H_{mc}^2$, *but there is no partial meet e-contraction yielding* $H'$.

In order to rectify this situation, we propose that *e*-contraction should be extended to include cases such as $H'$ above. The argument is as follows: Since full meet *e*-contraction is deemed to be appropriate, it stands to reason that any belief set $H'$ bigger than it should also be seen as appropriate, *provided* that $H'$ does not contain any irrelevant additions. But since $H'$ is contained in some maxichoice *e*-contraction, $H'$ cannot contain any irrelevant additions. After all, the maxichoice Horn *e*-contraction contains only relevant additions, since it is an appropriate form of contraction. Hence $H'$ is also an appropriate result of *e*-contraction.

In summary, *every* Horn belief set between full meet and some maxichoice *e*-contraction ought to be seen as an appropriate candidate for *e*-contraction. This is captured by the definition below.

**Definition 24 (Infra *e*-Remainder Sets)** *For Horn belief sets $H$ and $H'$, $H' \in H \downarrow_e \varphi$ if and only if there is some $H'' \in H \perp_e \varphi$ such that $(\bigcap H \perp_e \varphi) \subseteq H' \subseteq H''$. We refer to the elements of $H \downarrow_e \varphi$ as the* infra *e*-remainder sets *of $H$ with respect to $\varphi$.*

As with the case for full propositional logic (cf. Definition 9), *e*-remainder sets are also infra *e*-remainder sets, and so is the intersection of any set of *e*-remainder sets. Similarly, the intersection of any set of infra *e*-remainder sets is also an infra *e*-remainder set, and the set of infra *e*-remainder sets contains *all* Horn belief sets between some *e*-remainder set and the intersection of all *e*-remainder sets. As in the full propositional case, this explains why *e*-contraction is not defined as the intersection of infra *e*-remainder sets (cf. Definition 4).

**Definition 25 (Horn *e*-Contraction)** *Let $H$ be a Horn belief set and $\varphi \in \mathcal{L}_\mathsf{H}$ be a Horn formula. An infra e-selection function $\tau$ is a (partial) function from $\mathscr{P}(\mathscr{P}(\mathcal{L}_\mathsf{H}))$ to $\mathscr{P}(\mathcal{L}_\mathsf{H})$ such that*

- $\tau(H \downarrow_e \varphi) = H$ *whenever* $H \downarrow_e \varphi = \emptyset$; *and*
- $\tau(H \downarrow_e \varphi) \in H \downarrow_e \varphi$ *otherwise.*

*An e-contraction $-_\tau$ is an infra e-contraction if and only if $H -_\tau \varphi = \tau(H \downarrow_e \varphi)$.*

The results on how kernel contraction, partial meet contraction and infra contraction compare for the base case (kernel contraction and infra contraction are identical, while both are more general than partial meet contraction) invite the question whether similar results hold for Horn belief sets. Before we provide an answer to this, we need a suitable version of kernel contraction for Horn belief sets.

**Definition 26 (Horn kernel *e*-contraction)** *Given a Horn belief set $H$ and an incision function $\sigma$ for $H$, the* Horn kernel *e*-contraction *for $H$, abbreviated as the* kernel *e*-contraction *for $H$, is defined as $H \approx_\sigma^e \varphi =_{def} Cn_{\mathsf{HL}}(H -_\sigma \varphi)$, where $-_\sigma$ is the belief base kernel contraction for $\varphi$ obtained from $\sigma$.*

It turns out that infra *e*-contraction and kernel *e*-contraction coincide, as the following result shows.

**Theorem 8** *An e-contraction for a Horn belief set $H$ is an infra e-contraction for $H$ if and only if it is a kernel e-contraction for $H$.*

From Theorem 8 and Example 4 it follows that partial meet *e*-contraction is more restrictive than kernel *e*-contraction. When it comes to Horn belief sets, we therefore have exactly the same pattern as we have for belief bases: kernel contraction and infra contraction coincide, while both are strictly more permissive than partial meet contraction. Contrast this with the case for belief sets for full propositional logic where infra contraction, partial meet contraction and kernel contraction all coincide.

One conclusion to be drawn from this is that the restriction to the Horn case produces a curious hybrid between belief sets and belief bases for full propositional logic. On the one hand, Horn contraction deals with sets that are logically closed. But on the other hand, the results for Horn logic obtained in terms of construction methods are close to those obtained for belief base contraction.

Either way, the new results on belief base contraction prove to be quite useful in the investigation of contraction for Horn belief sets. We conclude this section by providing a representation result for Horn contraction inspired by the new results on belief base contraction in this paper.

**Theorem 9** *Every infra e-contraction satisfies $(K-1)$, $(K-2)$, $(K-4)$, $(K-5)$ and Core-retainment. Conversely, every e-contraction which satisfies $(K-1)$, $(K-2)$, $(K-4)$, $(K-5)$, and Core-retainment is an infra e-contraction.*

This result was inspired by Theorem 7 which shows that base kernel and base infra contraction coincide. Given that Core-retainment is used in characterizing base kernel contraction, Theorem 7 shows that there is a link between Core-retainment and base infra contraction, and raises the question of whether there is a link between Core-retainment and infra *e*-contraction. The answer, as we have seen in Theorem 9, is yes. This result provides more evidence for the hybrid nature of contraction for Horn belief sets. In this case the connection with base contraction is strengthened.

## 6. Related Work

To our knowledge, the first formal proposal of taking non-maximal remainder sets for a contraction following the AGM principles was made by Restall and Slaney (1995). The construction appears in a context very different from ours: they use a four-valued logic and show that by dropping the Recovery postulate, they obtain a representation result for partial-meet with non-maximal remainders. Note that there is no restriction as to the remainder sets containing a minimal core, as was made for infra remainders, and as a result there is no postulate associated with any kind of minimal change.

While there has been some work on *revision* for Horn formulas (Eiter & Gottlob, 1992; Liberatore, 2000; Langlois, Sloan, Szörényi, & Turán, 2008), it is only recently that attention has been paid to *contraction* for Horn logic. Apart from the work of Delgrande (2008) which has been discussed, there is some recent work on obtaining a semantic characterisation of Horn contraction using the system of spheres by Fotinopoulos and Papadopoulos (2009), and via epistemic entrenchment by Zhuang and Pagnucco (2010).

Billington et al. (1999) considered revision and contraction for defeasible logic which is quite different from Horn logic in many respects, but nevertheless has a rule-like flavour to it which has some similarity to Horn logic.

The present paper is an extension of the work by Booth et al. (2009) in which they show that infra $e$-contraction is captured precisely by the six AGM postulates for belief set contraction, except that Recovery is replaced by the following (weaker) postulate $(H -_e 6)$ together with the Failure postulate:

$(H -_e 6)$ If $\psi \in H \setminus (H - \varphi)$, then
there exists an $X$ such that $\bigcap(H \perp_e \varphi) \subseteq X \subseteq H$ and $X \not\models \varphi$, but $X \cup \{\psi\} \models \varphi$

$(H -_e 7)$ If $\models \varphi$, then $H -_e \varphi = H$ \hfill (Failure)

**Theorem 10 (Booth et al., 2009)** *Every infra $e$-contraction operator satisfies Postulates $(K-1)$–$(K-5)$, $(H -_e 6)$ and $(H -_e 7)$. Conversely, every $e$-contraction which satisfies $(K-1)$–$(K-5)$, $(H -_e 6)$ and $(H -_e 7)$ is an infra $e$-contraction.*

Postulate $(H -_e 6)$ bears some resemblance to the Relevance postulate for base contraction in that it states that all sentences removed from $H$ during a $\varphi$-contraction must have been removed for a reason: adding them again brings back $\varphi$. Postulate $(H -_e 7)$ simply states that contracting with a tautology leaves the initial belief set unchanged.

It is worth noting that $(H -_e 6)$ is a somewhat unusual postulate in that it refers directly to the construction method it is intended to characterize, i.e., to $e$-remainder sets. Here we have provided a more elegant characterization of infra $e$-contraction by taking a detour through base contraction. Firstly, we replaced $(H -_e 6)$ with the Core-retainment postulate that is used in the characterization of base kernel contraction. Then it turns out that the Vacuity postulate $(K-3)$ and the Failure postulate $(H -_e 7)$ both follow from Core-retainment and the Inclusion postulate $(K-2)$, from which we then obtained our characterization of infra $e$-contraction.

In addition to $e$-contraction, Delgrande also investigated a version of Horn contraction he refers to as *inconsistency-based contraction* (or $i$-contraction) where the purpose is to modify an agent's Horn belief set in such a way as to avoid inconsistency when a sentence $\varphi$ is provided as input. That is, an $i$-contraction $-_i$ should be such that $(H -_i \varphi) \cup \{\varphi\} \not\models_{\mathsf{HL}} \bot$.

In addition to both $e$- and $i$-contraction, Booth et al. (2009) considered *package contraction* (or $p$-contraction), a version of contraction studied by Fuhrmann and Hansson (1994) for the classical case (i.e., for logics containing full propositional logic). Given a set of formulas $X$, the goal is to make sure that *none* of the sentences in $X$ is in the result obtained from $p$-contraction. For full propositional logic this is similar to contracting with the disjunction of the sentences in $X$. For Horn logic, which does not have full disjunction, package contraction is more interesting. Although it seems that the new results presented in this paper can be applied to both $i$-contraction and $p$-contraction, this still has to be verified in detail.

Recent work by Delgrande and Wassermann (2010) draws inspiration from the semantic representation of remainder sets. In the classical AGM approach, a remainder set can be obtained semantically by adding to the models of a belief set $H$ a counter-model of the formula $\varphi$ for contraction. With Horn clauses, this construction will not necessarily lead to sets that correspond to remainder sets. This is because, as is known in Horn logic, given any two models of $m_1, m_2$ of a Horn belief set $H$, the model $m_1 \sqcap m_2$ is also a model of $H$, where $m_1 \sqcap m_2$ denotes that model in which the atoms true in it are precisely the atoms which are true in both $m_1$ and $m_2$. To illustrate this, consider the following example:

**Example 5 (Delgrande & Wassermann, 2010)** *Let $\mathfrak{P} = \{p, q, r\}$ and let the Horn belief set $H = Cn_{\mathsf{HL}}(p \wedge q)$. Consider candidates for $H - (p \wedge q)$. There are three e-remainder sets, given by the Horn closures of $p \wedge (r \to q)$, $q \wedge (r \to p)$, and $(p \to q) \wedge (q \to p) \wedge (r \to p) \wedge (r \to q)$. Any infra e-remainder set contains the closure of $(r \to p) \wedge (r \to q)$.*

In Example 5, the unique model of the set $\{p, \neg q, r\}$ is a countermodel of $p \wedge q$. If we add it to the models of $H$ and close them under $\sqcap$ so that the result represents a Horn belief set, we obtain the set of models of the formula $p$, which cannot be an e-remainder set, as there is an e-remainder set ($Cn_{\mathsf{HL}}(p \wedge (r \to q))$) containing it. This means that in an approach based on e-remainder sets we cannot have contraction operators with the same behaviour as the classical ones.

Delgrande and Wassermann propose to mimic the semantic construction of classical remainders: a *weak-remainder* of $H$ and $\varphi$ is formed by intersecting $H$ with a maximal consistent Horn theory not containing $\varphi$. Equivalently, the weak-remainders can be semantically characterized by taking the closure under intersection of the models of $H$ together with a single counter-model of $\varphi$. Representation results for contraction based on weak-remainders are then provided.

Concerning the connection between partial meet contraction and kernel contraction for bases, Falappa et al. (2006) have shown how to construct a partial meet contraction from a kernel contraction, which is always possible, and how to construct a kernel contraction from a partial meet contraction, when this is possible. These results can be generalized so that they apply to base infra contraction and base kernel contraction. That is, it should be possible to construct a base infra contraction from every base kernel contraction, as well as a base kernel contraction from every base infra contraction.

## 7. Concluding Remarks

In bringing Hansson's kernel contraction into the picture, we have made meaningful contributions to the investigation into contraction for Horn logic. The main contributions of the present paper are as follows: (*i*) A result which shows that infra contraction and kernel contraction for the belief base case coincide; and (*ii*) Lifting the previous results to Horn belief sets to show that infra contraction and kernel contraction for Horn belief sets coincide.

The investigation into base contraction also allowed us to improve on the rather unsatisfactory representation result proved by Booth et al. for infra contraction, which relies on a postulate referring directly to the construction method it is intended to characterize. We obtained a more elegant representation result by replacing the postulate introduced by Booth et al. with the well-known Core-retainment postulate, which is usually associated with base contraction. The presence of Core-retainment here further enforces the hybrid nature of Horn belief change, lying somewhere between belief set change and base change.

We have seen that kernel $e$-contraction and infra $e$-contraction are more general than partial meet $e$-contraction. But there is also evidence that even these forms of Horn contraction may not be sufficient to obtain all meaningful answers. Consider, for example, our Horn belief set example $Cn_{\mathsf{HL}}(\{p \to q, q \to r\})$ encountered in Example 4. If we view basic Horn clauses (clauses with exactly one atom in the head and the body) as representative of arcs in a graph, in the style of the old inheritance networks, there is a case to be made for regarding the three "arcs" $p \to q$, $q \to r$, and $p \to r$, as the *basic* information used to generate the belief set. Then one possible desirable outcome of a contraction by $p \to r$ is $Cn_{\mathsf{HL}}(q \to r)$. However, as we have seen in Example 4, this is not an outcome supported by infra $e$-contraction (and therefore not by kernel $e$-contraction either). Ideally, a truly comprehensive $e$-contraction approach for Horn logic would be able to account for such cases as well.

Here we focus only on basic Horn contraction. For future work we plan to investigate Horn contraction for *full* AGM contraction, obtained by adding the *extended* postulates. Another interesting question for further investigation is whether Theorem 9 can be generalized to other logics, notably extensions of (full) propositional logic.

Finally, as mentioned earlier, one of the reasons for focusing on this topic is because of its application to debugging and repairing ontologies in description logics. In particular, Horn logic can be seen as the backbone of the $\mathcal{EL}$ family of description logics (Baader et al., 2005), and therefore a proper understanding of belief change for Horn logic is important for finding solutions to similar problems expressible in the $\mathcal{EL}$ family. We are currently investigating these possibilities.

## Acknowledgments

This paper extends the work by Booth, Meyer and Varzinczak that appeared at IJCAI (Booth et al., 2009), and the work by Booth, Meyer, Varzinczak and Wassermann that appeared at NMR (Booth et al., 2010a) and at ECAI (Booth et al., 2010b).

The authors are grateful to the anonymous referees for their constructive and useful remarks, which helped improving the quality and presentation of this work. We would also like to thank Eduardo Fermé who provided important hints on the connection between infra and kernel contraction.

The work of Richard Booth was supported by the FNR INTER project "Dynamics of Argumentation". The work of Thomas Meyer and Ivan Varzinczak was supported by the National Research Foundation under Grant number 65152. The work of Renata Wassermann was supported by CNPq, the Brazilian National Research Council, under grants 304043/2010-9 and 471666/2010-6.

## Appendix A. Proofs of Main Theorems

For the proofs of Lemma A.1 and Theorem 3 we need a few model-theoretic notions. We denote by $\mathcal{W}$ the set of valuations of $\mathcal{L}_{\mathsf{P}}$ and by $[\varphi]$ the set of models of $\varphi$, i.e., the set of valuations which satisfy $\varphi$. For a set of sentences $X$, we denote by $[X]$ the set of models of $X$, i.e., the set of valuations which satisfy all sentences in $X$. For any $V \subseteq \mathcal{W}$, we let $Th(V) = \{\varphi \in \mathcal{L}_\mathsf{P} \mid V \subseteq [\varphi]\}$.

The results in Lemma A.1 below are well-known model-theoretic descriptions of full meet, partial meet, and maxichoice contraction, first presented by Katsuno & Mendelzon (1991). The results, as stated below, are not stated directly by Katsuno & Mendelzon, but follow as easy consequences of their Theorem 3.3.

**Lemma A.1 (Katsuno & Mendelzon, 1991)** *Let $K$ be a belief set.*

1. *Let $-_{fm}$ be the (belief set) full meet contraction. For every $\varphi \in \mathcal{L}_\mathsf{P}$, $K -_{fm} \varphi = Th([K] \cup [\neg\varphi])$;*

2. *Let $-_{mc}$ be a (belief set) maxichoice contraction. For every $\varphi \in \mathcal{L}_\mathsf{P}$, there is a $w \in [\neg\varphi]$ such that $K -_{mc} \varphi = Th([K] \cup \{w\})$;*

3. *Given $\varphi \in \mathcal{L}_\mathsf{P}$, let $V_\varphi \subseteq [\neg\varphi]$ be such that $V_\varphi \neq \emptyset$. There is a (belief set) partial meet contraction $-_{pm}$ such that for every $\varphi \in \mathcal{L}_\mathsf{P}$, $K -_{pm} \varphi = Th([K] \cup V_\varphi)$.*

**Theorem 3 (Convexity)** *Let $K$ be a belief set, let $-_{mc}$ be a (belief set) maxichoice contraction, and let $-_{fm}$ denote (belief set) full meet contraction. For every $\varphi \in \mathcal{L}_\mathsf{P}$, let $X_\varphi$ be a belief set such that $(K -_{fm} \varphi) \subseteq X_\varphi \subseteq K -_{mc} \varphi$. There is a (belief set) partial meet contraction $-_{pm}$ such that for every $\varphi \in \mathcal{L}_\mathsf{P}$, $K -_{pm} \varphi = X_\varphi$.*

**Proof:**
The cases where $\models \varphi$ and $\varphi \notin K$ hold easily, so we suppose that $\not\models \varphi$ and $\varphi \in K$. Now, pick any belief set $X_\varphi$ such that $K -_{fm} \varphi \subseteq X_\varphi \subseteq K -_{mc} \varphi$. By Points (1) and (2) of Lemma A.1 this means that there is a $w \in [\neg\varphi]$ such that $[K] \cup \{w\} \subseteq [X_\varphi] \subseteq K \cup [\neg\varphi]$. From this it follows that $[X_\varphi] = [K] \cup [V_\varphi]$ for some $V_\varphi \subseteq [\neg\varphi]$. And from Point (3) of Lemma A.1 it then follows that there is a partial meet contraction $-_{pm}$ such that, for every $\varphi \in \mathcal{L}_\mathsf{P}$, $K -_{pm} \varphi = X_\varphi = Th([K] \cup V_\varphi)$. [QED]

**Lemma A.2 (Falappa et al., 2006)** *Every base kernel contraction defined by a minimal incision function is a base maxichoice contraction.*

**Lemma A.3** *The base kernel contraction defined by the maximal incision function is the full meet base contraction.*

**Proof:**
We need to prove that $B \setminus \bigcup(B \bot\!\!\!\bot \varphi) = \bigcap(B\bot\varphi)$. If $\models \varphi$ or $B \not\models \varphi$ the result follows immediately. So we suppose that $\not\models \varphi$ and $B \models \varphi$. For the left-to-right direction, suppose that $\psi \in B$ and $\psi \notin \bigcup(B \bot\!\!\!\bot \varphi)$, and assume there is some base remainder set of $B$ with respect to $\varphi$, say $X$, such that $\psi \notin X$. Then it must be the case that $\psi$ is contained in some $\varphi$-kernel of $B$. But then $\psi \in \bigcup(B \bot\!\!\!\bot \varphi)$: a contradiction. For the right-to-left direction, suppose $\psi$ is in every base remainder set of $B$ with respect to $\varphi$, and assume that $\psi \notin B \setminus \bigcup(B \bot\!\!\!\bot \varphi)$. Since $\psi \in B$ it follows that $\psi \in \bigcup(B \bot\!\!\!\bot \varphi)$. That is, $\psi$ is in some $\varphi$-kernel of $B$. So there is at least one minimal incision function $\sigma$ such that $\psi \in \sigma(B \bot\!\!\!\bot \varphi)$.

Now, from Lemma A.2 it follows that there is a base remainder set of $B$ with respect to $\varphi$, say $Y$, such that $Y = B \setminus \sigma(B \!\perp\!\!\!\perp\! \varphi)$. So $\psi \notin Y$, which contradicts the supposition that $\psi$ is in every base remainder set of $B$ with respect to $\varphi$. [QED]

**Theorem 7** *A base contraction for a belief base $B$ is a belief base kernel contraction for $B$ if and only if it is a base infra contraction for $B$.*

**Proof:**
For the "only if" part, let $-$ be a base kernel contraction for $B$, let $\sigma$ be the incision function for $B$ which generates $-$, and pick a $\varphi \in \mathcal{L}_\mathsf{P}$. If $\models \varphi$ or $B \not\models \varphi$ the result holds easily, so we suppose that $\not\models \varphi$ and $B \models \varphi$. It remains to be shown that $\bigcap(B\!\perp\!\varphi) \subseteq B - \varphi \subseteq B'$ for some $B' \in B\!\perp\!\varphi$. Observe firstly that $B \setminus \bigcup(B \!\perp\!\!\!\perp\! \varphi) \subseteq B \setminus \sigma(B \!\perp\!\!\!\perp\! \varphi) = B - \varphi$. From Lemma A.3 it then follows directly that $\bigcap(B\!\perp\!\varphi) \subseteq B - \varphi$. Next, pick any minimal incision function $\sigma'$ such that $\sigma'(B \!\perp\!\!\!\perp\! \varphi) \subseteq \sigma(B \!\perp\!\!\!\perp\! \varphi)$ ($\sigma'$ clearly exists). So $B - \varphi = B \setminus \sigma(B \!\perp\!\!\!\perp\! \varphi) \subseteq B \setminus \sigma'(B \!\perp\!\!\!\perp\! \varphi)$ and from Lemma A.2 it follows that $B \setminus \sigma'(B \!\perp\!\!\!\perp\! \varphi) = B'$ for some $B' \in B\!\perp\!\varphi$.

For the "if" part, let $-$ be a base infra contraction for $B$. We construct an incision function $\sigma$ such that the base kernel contraction it generates is exactly $-$. Pick a $\varphi \in \mathcal{L}_\mathsf{P}$ and suppose that $\not\models \varphi$ and $B \models \varphi$ (for the remaining cases the result easily holds). Let $\sigma(B \!\perp\!\!\!\perp\! \varphi) = B \setminus (B - \varphi)$. Since $B - \varphi \subseteq B$ we have that $B - \varphi = B \setminus \sigma(B \!\perp\!\!\!\perp\! \varphi)$. First we need to show that $\sigma(B \!\perp\!\!\!\perp\! \varphi) \subseteq \bigcup(B \!\perp\!\!\!\perp\! \varphi)$. To do so, observe that since $B - \varphi \in B\!\downarrow\!\varphi$, it follows that $\bigcap(B\!\perp\!\varphi) \subseteq B - \varphi$. Therefore $B \setminus (B - \varphi) \subseteq B \setminus \bigcap(B\!\perp\!\varphi)$. By Lemma A.3 and the construction of $\sigma$ we then have that $\sigma(B \!\perp\!\!\!\perp\! \varphi) = B \setminus (B - \varphi) \subseteq B \setminus \bigcap(B\!\perp\!\varphi) = \bigcup(B \!\perp\!\!\!\perp\! \varphi)$.

It remains to be shown that for every $Y$ such that $\emptyset \subset Y \in B \!\perp\!\!\!\perp\! \varphi$, $Y \cap \sigma(B \!\perp\!\!\!\perp\! \varphi) \neq \emptyset$. Pick a $Y$ such that $\emptyset \subset Y \in B \!\perp\!\!\!\perp\! \varphi$, and assume that $Y \cap \sigma(B \!\perp\!\!\!\perp\! \varphi) = \emptyset$. Then it must be the case that $Y \subseteq B - \varphi$. But $Y \models \phi$ and therefore $B - \varphi \models \varphi$, a contradiction. [QED]

**Theorem 8** *An $e$-contraction for a Horn belief set $H$ is an infra $e$-contraction for $H$ if and only if it is a kernel $e$-contraction for $H$.*

**Proof:**
Consider a belief base $B$ and a formula $\varphi$. From Theorem 7 it follows that the set of base infra remainder sets of $B$ with respect to $\varphi$ (i.e., the set $B\!\downarrow\!\varphi$) is equal to the set of results obtained from the base kernel contraction of $B$ by $\varphi$, call it $KC_\varphi^B$. Now let $B$ be such that it is a set of Horn formulas closed under Horn consequence (a Horn belief set) and $\varphi$ a Horn formula. The elements of $KC_\varphi^B$ are not necessarily closed under Horn consequence, but if we do close them, we obtain exactly the set of results obtained from kernel $e$-contraction when contracting $B$ by $\varphi$ (by the definition of kernel $e$-contraction). Let us refer to this latter set as $Cn_\mathsf{HL}(KC_\varphi^B)$, i.e., $Cn_\mathsf{HL}(KC_\varphi^B) = \{Cn_\mathsf{HL}(X) \mid X \in KC_\varphi^B\}$. Also, the elements of $B \downarrow \varphi$ are not closed under Horn consequence, but if we do close them the resulting set (refer to this set as $Cn_\mathsf{HL}(B\!\downarrow\!\varphi)$) contains exactly the infra $e$-remainder sets of $B$ with respect to $\varphi$, i.e., $Cn_\mathsf{HL}(B\!\downarrow\!\varphi) = B\!\downarrow_e\!\varphi$. To see why, observe that since $B$ is closed under Horn consequence, the (base) remainder sets of $B$ with respect to $\varphi$ (i.e., the elements of $B\!\perp\!\varphi$) are also closed under Horn consequence. So the elements of $B\!\downarrow\!\varphi$ are all the sets (not necessarily closed under Horn consequence) between $\bigcap(B\!\perp\!\varphi)$ and some element of $B\!\perp\!\varphi$. Therefore the elements of $Cn_\mathsf{HL}(B\!\downarrow\!\varphi)$ are all those elements of $B\!\downarrow\!\varphi$ that are closed under Horn consequence, i.e., $Cn_\mathsf{HL}(B\!\downarrow\!\varphi) = B\!\downarrow_e\!\varphi$ as was claimed. But since $B\!\downarrow\!\varphi = KC_\varphi^B$, it is also the case that $Cn_\mathsf{HL}(B\!\downarrow\!\varphi) = Cn_\mathsf{HL}(KC_\varphi^B)$, and therefore $Cn_\mathsf{HL}(KC_\varphi^B) = B\!\downarrow_e\!\varphi$. [QED]

**Theorem 9** *Every infra e-contraction satisfies* $(K-1)$, $(K-2)$, $(K-4)$, $(K-5)$ *and Core-retainment. Conversely, every e-contraction which satisfies* $(K-1)$, $(K-2)$, $(K-4)$, $(K-5)$, *and Core-retainment is an infra e-contraction.*

**Proof:**
Let $H$ be a Horn belief set, and let $-$ be an infra e-contraction. If $H - \varphi = H$, then $(K-1)$, $(K-2)$, $(K-4)$, $(K-5)$ and Core-retainment are trivially satisfied. Suppose $H - \varphi \neq H$. Then $H - \varphi = X$ for some $X \in H \downarrow_e \varphi$. Then, by definition, $H - \varphi$ is a Horn belief set, and hence $H - \varphi = Cn_{\mathsf{HL}}(H - \varphi)$ (Postulate $(K-1)$). Moreover, there is an $X' \in H \perp_e \varphi$ such that $X \subseteq X'$. Since $X' \subseteq H$, it also follows that $X \subseteq H$, and then $H - \varphi \subseteq H$ (Postulate $(K-2)$). $(K-4)$ is also satisfied from the definition of $H \downarrow_e \varphi$ and the monotonicity of classical logic. $(K-5)$ follows straightforwardly from the fact that we are working with Horn belief sets. Finally, for Core-retainment, suppose that $\psi \in H$ and $\psi \notin H - \varphi$. Now assume that $H' \cup \{\psi\} \not\models_{\overline{\mathsf{HL}}} \varphi$ for every $H' \subseteq H$ such that $H' \not\models_{\overline{\mathsf{HL}}} \varphi$. In particular then, for every $H' \in H \perp_e \varphi$, $H' \cup \{\psi\} \not\models_{\overline{\mathsf{HL}}} \varphi$. From that it follows that $\psi \in H'$ for every $H' \in H \perp_e \varphi$, and therefore that $\psi \in \bigcap(H \perp_e \varphi)$. And from this we have that $\psi \in H - \varphi$, which contradicts our supposition.

Conversely, let $H$ be a Horn belief set and let $-$ be an e-contraction which satisfies Postulate $(K-1)$, $(K-2)$, $(K-4)$, $(K-5)$, and Core-retainment. We need to show that for every $\varphi \in \mathcal{L}_{\mathsf{H}}$, $H - \varphi \in H \downarrow_e \varphi$. Observe firstly that $-$ satisfies $(K-3)$. To see why, note that from Core-retainment it follows that if $\varphi \notin H$ then $H \setminus (H - \varphi) = \emptyset$, and therefore that $H \subseteq H - \varphi$. From $(K-2)$ it then follows that $H = H - \varphi$.

If $\varphi \notin H$ it follows directly from $(K-3)$ that $H - \varphi \in H \downarrow_e \varphi$. If $\models_{\overline{\mathsf{HL}}} \varphi$ it follows directly from $(K-2)$ and Core-retainment that $H - \varphi = H$, and therefore that $H - \varphi \in H \downarrow_e \varphi$.

So, we suppose that $\varphi \in H$ and $\not\models_{\overline{\mathsf{HL}}} \varphi$. To show that $H - \varphi \in H \downarrow_e \varphi$, we need to show that $H - \varphi$ is a Horn belief set (from Postulate $(K-1)$) such that $\bigcap(H \perp_e \varphi) \subseteq H - \varphi \subseteq H'$ for some $H' \in H \perp_e \varphi$. Observe firstly that, since $H - \varphi \subseteq H$ (by Postulate $(K-2)$) and $\varphi \notin H - \varphi$ (by $(K-4)$), there has to be an $H' \in H \perp_e \varphi$ such that $H - \varphi \subseteq H'$. Finally, assume there is a $\psi \in \bigcap(H \perp_e \varphi)$ such that $\psi \notin H - \varphi$. By Core-retainment it then follows that there is an $H'' \subseteq H$ such that $H'' \not\models_{\overline{\mathsf{HL}}} \varphi$ but $H'' \cup \{\psi\} \models_{\overline{\mathsf{HL}}} \varphi$. From this it follows that there is a Horn belief set $X$ such that $H'' \subseteq X$ and $X \not\models_{\overline{\mathsf{HL}}} \varphi$ but $X \cup \{\psi\} \models_{\overline{\mathsf{HL}}} \varphi$. So $X \in H \perp_e \varphi$, but $\psi \notin X$, which contradicts our assumption that $\psi \in \bigcap(H \perp_e \varphi)$. It therefore follows that $\bigcap(H \perp \varphi) \subseteq H - \varphi$. [QED]